\documentclass[11pt,a4paper,english]{article}
\usepackage[utf8]{inputenc}
\usepackage{array}
\usepackage{multirow}
\usepackage{amsmath}
\usepackage{amssymb}
\usepackage{graphicx}

\makeatletter

\pdfpageheight\paperheight
\pdfpagewidth\paperwidth

\providecommand{\tabularnewline}{\\}

\usepackage[hyperref]{acl2020}
\usepackage{times}
\usepackage{latexsym}

\aclfinalcopy

\makeatother

\usepackage{babel}
\begin{document}
\title{Learning to Detect Unacceptable Machine Translations\\
for Downstream Tasks}
\author{Meng Zhang$^\dagger$ Xin Jiang$^\dagger$ Yang Liu$^\ddagger$ Qun Liu$^\dagger$\\
$^\dagger$Huawei Noah's Ark Lab\\
$^\ddagger$Institute for Artificial Intelligence\\
State Key Laboratory of Intelligent Technology and Systems\\
Department of Computer Science and Technology, Tsinghua University, Beijing, China\\
Beijing National Research Center for Information Science and Technology\\
$^\dagger$\texttt{\{zhangmeng92, Jiang.Xin, qun.liu\}@huawei.com}\\
$^\ddagger$\texttt{liuyang2011@tsinghua.edu.cn}}
\maketitle
\begin{abstract}
The field of machine translation has progressed tremendously in recent
years. Even though the translation quality has improved significantly,
current systems are still unable to produce uniformly acceptable machine
translations for the variety of possible use cases. In this work,
we put machine translation in a cross-lingual pipeline and introduce
downstream tasks to define task-specific acceptability of machine
translations. This allows us to leverage parallel data to automatically
generate acceptability annotations on a large scale, which in turn
help to learn acceptability detectors for the downstream tasks. We
conduct experiments to demonstrate the effectiveness of our framework
for a range of downstream tasks and translation models.
\end{abstract}

\section{Introduction\label{sec:Introduction}}

The past few years have witnessed strong performance gains for machine
translation (MT), especially since the rise of neural machine translation
\cite{sutskever_sequence_2014,bahdanau_neural_2015}. Some recent
research reports that the neural machine translation results are comparable
to those by human translators in certain domains \cite{wu_googles_2016,hassan_achieving_2018}.

However, such success comes with several conditions, most notably
a large parallel corpus for training and a close domain for testing.
Even with plentiful resources, state-of-the-art systems like Google
Translate may still produce erroneous translations that baffle users
\cite{zheng_testing_2018}. In a word, a general-purpose machine translation
system that consistently produces human-level translations in open
domains is still yet to come. Therefore, in many real-world applications,
it would be desirable to predict whether the MT output fits the purpose.

\begin{figure}
\begin{centering}
\includegraphics{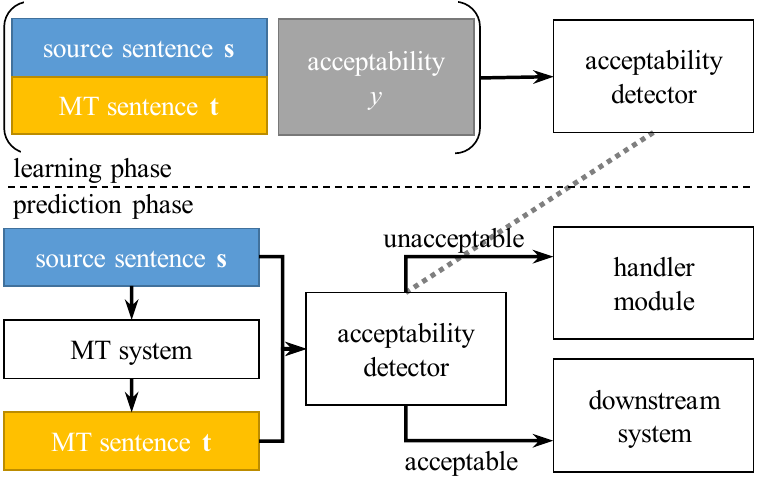}
\par\end{centering}
\caption{\label{fig:summary}A schematic illustration of our approach. In the
learning phase, automatically generated acceptability labels (Section
\ref{sec:Automatic-Acceptability-Annotation}) are used to train the
acceptability detector, which is used in the prediction phase to judge
whether the MT sentence is acceptable for the downstream system.}
\end{figure}

As an attempt to address the above issue, researchers propose the
machine translation quality estimation (QE) task \cite{specia_estimating_2009},
aiming to predict the quality of the translated text without access
to the reference text. The typical sentence-level QE task is framed
as supervised regression towards HTER \cite{snover_study_2006}, a
quality score between 0 and 1 defined with respect to the human post-edited
text. However, such a problem formulation brings several limitations:
\begin{itemize}
\item Gathering data labeled with HTER requires human post-editing effort,
which is costly to collect.
\item By definition, HTER is only informative for the scenario where MT
is used for post-editing.
\item HTER, as a single real-value score, is difficult for users to interpret
its exact meaning for the quality of machine translation \cite{turchi_data-driven_2014}.
\end{itemize}
A key motivation in this work is that MT can be used in various scenarios,
and their quality needs may differ. For example, fields like law or
patent have high quality standards that go beyond adequacy and fluency
towards style, while for purposes like gisting, loss of unimportant
information in the source text or small grammatical errors in the
target text can be tolerated.

\begin{figure*}
\begin{centering}
\includegraphics[scale=0.9]{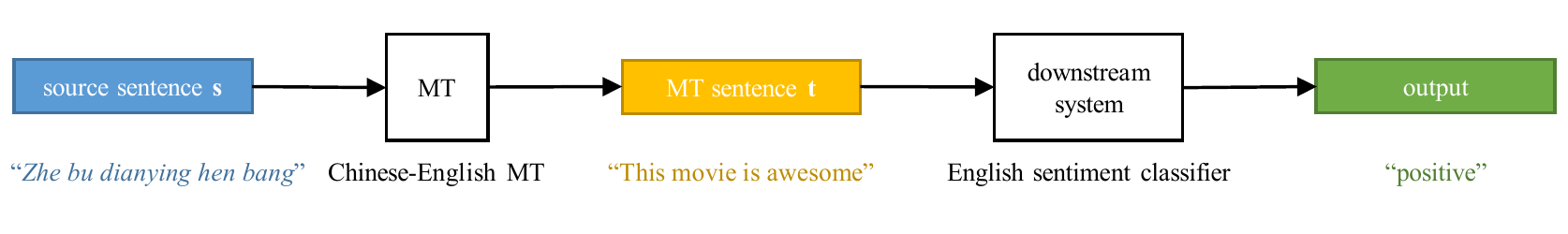}
\par\end{centering}
\caption{\label{fig:pipeline}The cross-lingual pipeline that first translates
the source sentence and then feeds the translation into a downstream
system in the target language. An illustrative instance is given below
the boxes for Chinese-English cross-lingual sentiment classification.}
\end{figure*}

We focus our attention on the usage scenarios where MT systems supply
output to downstream tasks that can be executed without human involvement,
and form an automatic cross-lingual pipeline (Figure \ref{fig:pipeline}).
This pipeline system is useful when the downstream task of interest
has automatic tools in the target language, but not in the source
language \cite{klementiev_inducing_2012,zhou_cross-lingual_2016,araujo_evaluation_2016}.
However, its usability clearly depends on the quality of MT, and our
goal is to build an automatic quality control system for MT to enhance
this pipeline.

In our setting, the downstream systems specify the MT quality standards
for their corresponding own tasks, so we introduce a new notion for
task-specific MT quality called \emph{acceptability} (Section \ref{sec:Acceptability}).
Since we can run the downstream systems automatically, it allows us
to leverage the large-scale parallel data and generate the labels
of acceptability (Section \ref{sec:Automatic-Acceptability-Annotation}),
which are used to supervise the learning of our acceptability detection
models (Section \ref{sec:Acceptability-Detection-Models}). As the
quality labels are produced specifically for each downstream task,
the same type of learning model can automatically adapt to the need
of the corresponding tasks during training. The trained acceptability
detector can then be integrated into the cross-lingual pipeline to
perform quality control for MT. Figure \ref{fig:summary} provides
an illustration for the process.

Our experiments demonstrate the advantages of our framework: the adaptability
to different downstream tasks, the benefit of automatic generation
of large-scale quality labels, and the applicability to different
translation models.

\section{Acceptability: Binary Task-Specific Machine Translation Quality\label{sec:Acceptability}}

As introduced in Figure \ref{fig:pipeline}, we consider the cross-language
processing problem where we are interested in performing a task on
the source language text, for example sentiment classification on
Chinese, but the NLP system for this task is only available in the
target language (e.g., English). A simple and general solution to
cross the language barrier is to make use of a machine translation
system to translate the source language text into the target language,
and then perform the task with the NLP system in the target language.
The solution is general in the sense that the MT system is agnostic
to the downstream task and is supposed to accommodate every possible
need. In particular, it should preserve all information of the source
language sentence (so that the downstream system can extract the relevant
information), and render the information in a fluent target language
sentence (because the downstream system is only trained in such condition).
Downstream tasks are many, with common ones including sentiment analysis,
spam detection, intent classification, and named entity recognition.
Naturally, the tasks may vary dramatically in nature, while a one-size-fits-all
MT is expected to meet the quality needs for all of them.

\begin{figure*}
\begin{centering}
\includegraphics{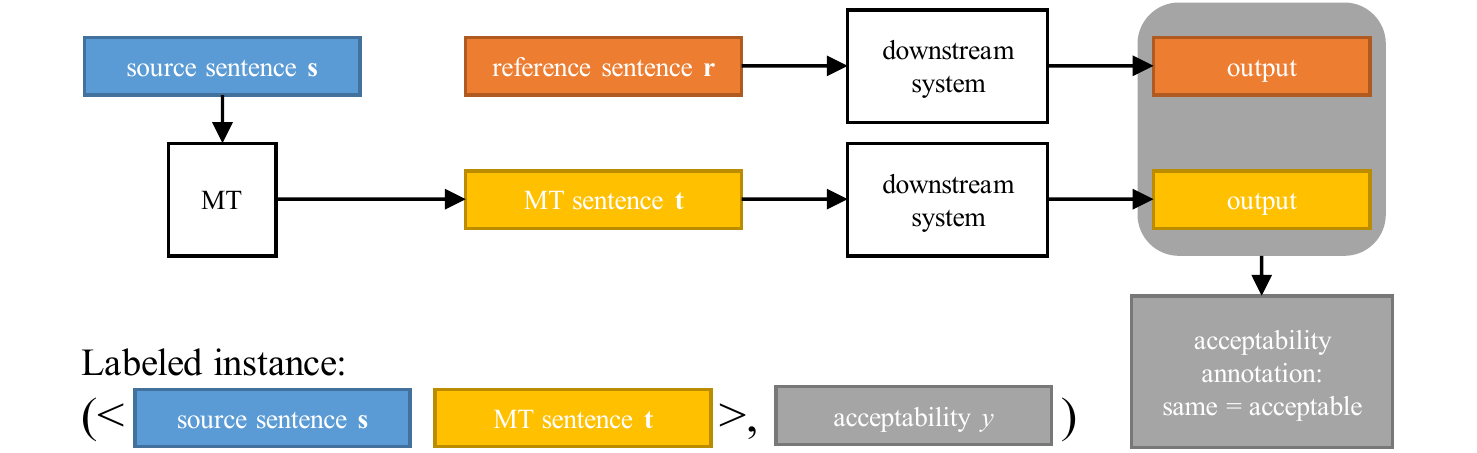}
\par\end{centering}
\caption{\label{fig:annotation}The process of using a parallel sentence pair
and a downstream system to automatically annotate acceptability to
obtain a labeled instance.}
\end{figure*}

Therefore, we alleviate the burden of the MT system by relaxing the
quality expectation. With the involvement of downstream system, we
propose that a machine translated text meets the quality need as long
as it suffices to make the downstream system function properly. Thereby,
we introduce a new notion of machine translation quality specific
to the downstream task, which we call \emph{acceptability} and define
it as follows.
\begin{description}
\item [{Definition}] (\emph{acceptability}): A machine translated text
is acceptable for a given downstream task if and only if the information
required for the downstream task is correctly transferred from the
source text.
\end{description}
For instance, given sentiment analysis as the downstream task, the
translation is acceptable provided that the translated text has the
same polarity as the source text.

More formally, let ${\bf s}$ be a source sentence, ${\bf t}={\rm MT}\left({\bf s}\right)$
be its machine translation. A program that executes the downstream
task and returns the result of interest is denoted as a function $f$,
with $f^{{\rm S}}$ and $f^{{\rm T}}$ acting on the source and target
language, respectively. Then the definition of acceptability can be
written as
\begin{equation}
\begin{aligned} & {\rm acceptability}_{f}\left({\bf s},{\bf t}\right)={\rm true}\\
 & \iff f^{{\rm S}}\left({\bf s}\right)=f^{{\rm T}}\left({\bf t}\right).
\end{aligned}
\end{equation}

Note the dependence of the acceptability on $f$, which reflects the
task-specific nature of the translation quality. This means a piece
of translation unacceptable for one task may be well acceptable for
another, and vice versa.

Also note that, conceptually, an ideal semantic extraction function
$f$ would define what humans perceive as ``acceptable translation''.
However, for practical purpose, we restrict our attention to functions
implemented by automatic downstream systems. This means acceptability
defined with respect to such functions does not necessarily agree
with human perception of translation quality.

In our problem setting, $f^{{\rm S}}$ is unavailable, which is the
reason we introduce MT at first place. To make the definition workable,
we introduce the reference translation ${\bf r}$ of the source sentence
${\bf s}$, giving

\begin{equation}
\begin{aligned} & {\rm acceptability}_{f}\left({\bf s},{\bf t}\right)={\rm true}\\
 & \iff f^{{\rm T}}\left({\bf t}\right)=f^{{\rm T}}\left({\bf r}\right).
\end{aligned}
\label{eq:2}
\end{equation}

In practice, the actual acceptability may be compromised by several
factors. For example, the reference translation may be noisy, or introduce
cultural difference across languages \cite{mohammad_how_2016}, or
obfuscate certain source text traits like author's gender \cite{mirkin_motivating_2015,rabinovich_personalized_2017},
so we need to make the assumption that the reference translation preserves
the information needed by the downstream task. For another, automatic
downstream systems that implement $f^{{\rm T}}$ are almost always
imperfect. In this regard, we assume that the downstream systems are
reliable enough, otherwise it would be inconceivable to achieve desired
results with the cross-lingual pipeline.

In the next section, we show how to automatically collect the annotations
of acceptability for machine translation.

\section{Automatic Acceptability Annotation\label{sec:Automatic-Acceptability-Annotation}}

To obtain training instances for the acceptability detection system,
instead of annotating acceptability manually, we can automate the
process with existing parallel sentence pairs by virtue of the machine
executability of $f^{{\rm T}}$. This process is illustrated in Figure
\ref{fig:annotation}.

The MT system in Figure \ref{fig:annotation} is the one used in the
cross-lingual pipeline, and its quality is what we care about. First
we use the MT system to translate every source sentence ${\bf s}$
in the parallel corpus. Then the translated sentence ${\bf t}$ is
paired with its reference ${\bf r}$ and fed into the downstream system
$f^{{\rm T}}$ to obtain their respective outputs. Finally the outputs
of each pair are compared to generate the acceptability annotation,
which is denoted as $y$. When the process is complete, we gather
tuples of $\left(\left\langle {\bf s},{\bf t}\right\rangle ,y\right)$
as the training instances of the acceptability detection system.

The acceptability detector is trained to predict acceptability $y$
given $\left\langle {\bf s},{\bf t}\right\rangle $ as input. Once
trained, it can be incorporated into the cross-lingual pipeline as
a new component, as shown in the lower part of Figure \ref{fig:summary}.
The handler module that takes unacceptable MT sentences may perform
actions specific to each task, and the most general way of handling
is probably presenting the case to human. However, if the downstream
task is binary classification, we can still feed the unacceptable
MT sentence into the downstream system and then flip the predicted
label.

We elaborate the learning of the acceptability detection models in
the next section.

\begin{figure*}
\begin{centering}
\includegraphics{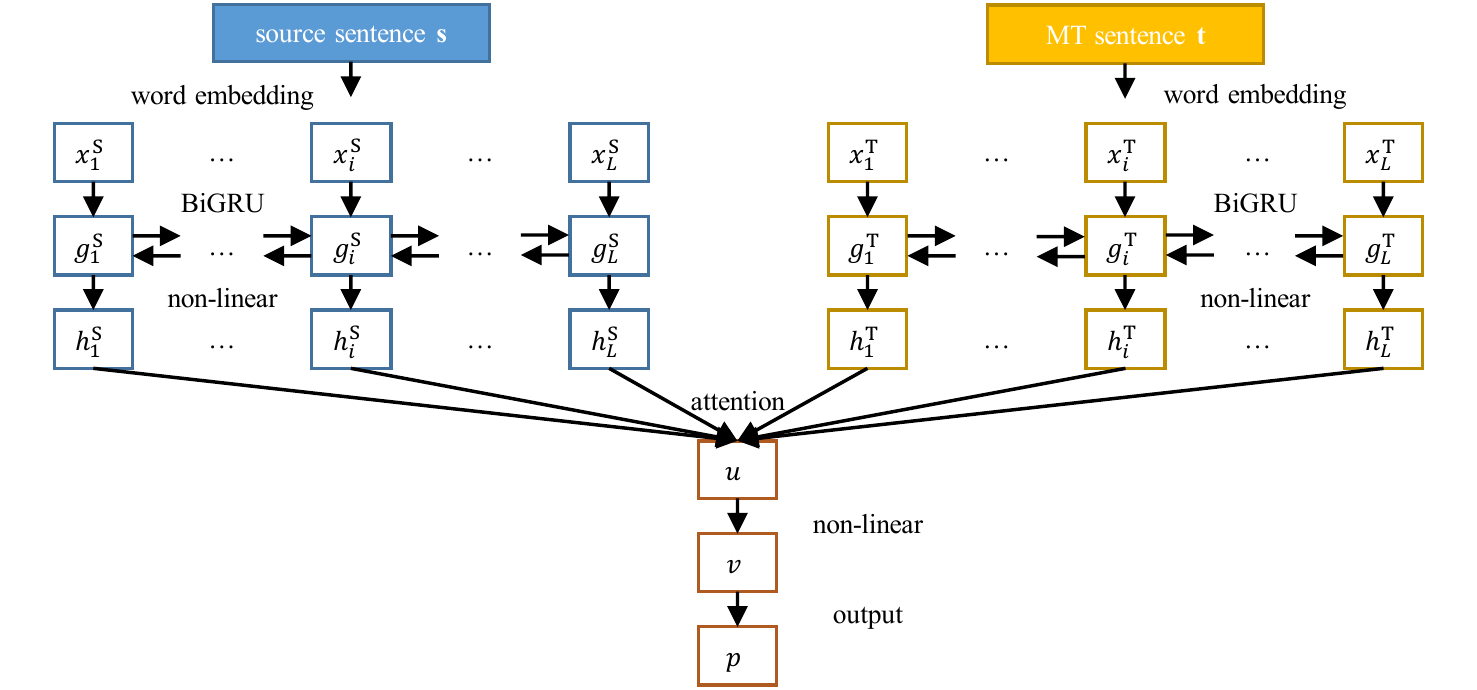}
\par\end{centering}
\caption{\label{fig:BiRNN}The BiRNN architecture.}
\end{figure*}

\section{Acceptability Detection Models\label{sec:Acceptability-Detection-Models}}

After we have gathered instances of $\left(\left\langle {\bf s},{\bf t}\right\rangle ,y\right)$,
we are ready to build the acceptability detector: $y=A_{f}\left({\bf s},{\bf t};\theta\right)$.
From the machine learning point of view, the difference between acceptability
detection and quality estimation is the type of the target variable
$y$, which leads to the learning problem of acceptability detector
being framed as binary classification. This connection hints that
models developed in the quality estimation literature may also be
effective for acceptability detection. Therefore, we draw inspiration
from the existing quality estimation methods and propose to approach
acceptability detection with two different models, called BiQuEst
and BiRNN respectively.

\subsection{BiQuEst}

QuEst \cite{specia_quest_2013} is a traditional method for quality
estimation, serving as the official baseline of the quality estimation
shared task since 2012 \cite{callison-burch_findings_2012}. It works
by extracting various human-designed features of $\left\langle {\bf s},{\bf t}\right\rangle $
that could be indicative of its quality. We hope these features are
also helpful for our acceptability detection purpose, and use them
to represent instances of $\left\langle {\bf s},{\bf t}\right\rangle $.
In our experiments, we use the default 17 blackbox features\footnote{https://www.quest.dcs.shef.ac.uk/quest\_files\slash features\_blackbox\_baseline\_17}
extracted by the QuEst++ toolkit \cite{specia_multi-level_2015}.
We then extend it to learn from binary labels (hence the name BiQuEst)
by using the support vector machine.

\subsection{BiRNN}

One difference between acceptability detection and quality estimation
is that the same $\left\langle {\bf s},{\bf t}\right\rangle $ pair
may receive different acceptability labels depending on the downstream
task. However, the features extracted by QuEst for the same $\left\langle {\bf s},{\bf t}\right\rangle $
are fixed, and the different labels can only be reflected by the weights
of the model. This limitation can be overcome by an end-to-end neural
network model that performs feature extraction and classification
as a whole.

We propose BiRNN as illustrated in Figure \ref{fig:BiRNN}. This model
is adapted from \cite{ive_deepquest:_2018}. The source and MT sentences
${\bf s}$ and ${\bf t}$ are represented as sequences of vectors
by separate word embedding layers, bidirectional GRU \cite{bahdanau_neural_2015}
layers, and non-linear layers:
\begin{align}
x_{i}^{{\rm S}} & =W_{e}^{{\rm S}}s_{i},i\in\left[1,L\right],\\
g_{i}^{{\rm S}} & ={\rm BiGRU}^{{\rm S}}\left(x_{i}^{{\rm S}}\right),i\in\left[1,L\right],\\
h_{i}^{{\rm S}} & ={\rm ReLU}\left(W_{g}^{{\rm S}}g_{i}^{{\rm S}}+b_{g}^{{\rm S}}\right),i\in\left[1,L\right].
\end{align}
The target language part shares similar equations. The two sentence
vectors $\left\{ h_{i}^{{\rm S}}\right\} _{i=1}^{L}$ and $\left\{ h_{i}^{{\rm T}}\right\} _{i=1}^{L}$
are concatenated together as a sequence of length $2L$, denoted as
$\left\{ h_{i}\right\} _{i=1}^{2L}$. They are then combined into
a single vector with the word attention mechanism:
\begin{align}
\alpha_{i} & =\frac{\exp\left(h_{i}^{\top}w\right)}{\sum_{j=1}^{2L}\exp\left(h_{j}^{\top}w\right)},i\in\left[1,2L\right],\\
u & =\sum_{i=1}^{2L}\alpha_{i}h_{i}.
\end{align}
Finally, this vector is passed through a non-linear layer before it
is used to compute the probability of the acceptability:
\begin{align}
v & ={\rm ReLU}\left(W_{u}u+b_{u}\right),\\
p & ={\rm sigmoid}\left(W_{v}v+b_{v}\right).
\end{align}
The cross-entropy loss is used for learning.

\section{Experimental Setup}

A direct way to demonstrate the efficacy of the acceptability detector
would be evaluating the performance gain from plugging it into the
cross-lingual pipeline. However, the task-specific handler module
that generally involves human complicates the creation of a uniform
test set. Even for binary classification downstream tasks, a suitable
cross-lingual data set is difficult to obtain. Therefore, we decide
to use the parallel data with automatic acceptability annotations
to create the experiment suite.

We take Chinese as the source language and English as the target,
and use parallel data from WMT18 \cite{bojar_findings_2018}. After
generating acceptability annotations, we reserve 10k instances as
the development set and another 10k instances as the test set. In
addition to acceptability detection models BiQuEst and BiRNN, we also
report the performance of a baseline that takes every translation
as acceptable, which corresponds to the original cross-lingual pipeline
in Figure \ref{fig:pipeline}.

We lowercase the English side of all parallel corpora. The MT training
corpus includes LDC2002E18, LDC2003E07, LDC2003E14, Hansards portion
of LDC2004T07, LDC2004T08, and LDC2005T06. For training both Transformer
and Moses translation models, we apply byte pair encoding \cite{sennrich_neural_2016}
with 32,000 operations. Note that when we use the 0.1m LDC subset,
the coding scheme differs accordingly. When we generate acceptability
annotation using the WMT18 corpus, only the source sentences with
50 subwords or fewer are retained. The BiRNN model takes input at
the subword level with maximum sequence length $L=64$, so no source
sentence is truncated.

All existing toolkits follow default settings. The Moses uses a 3-order
language model trained on the target side of the parallel corpus,
which is correspondingly smaller when the 0.1m LDC subset is used.
The Stanford NER distinguishes three types: person, location, and
organization.

The model hyperparameters are set as follows. For BiQuEst, 3-order
language models are trained on the 1.25m LDC parallel corpus for each
language to be used for feature extraction. For SVM learning, we use
the RBF kernel as it shows slightly better performance on the development
set than the linear kernel, although the training time is much longer.
The other hyperparameters are left as default. For BiRNN, the vocabulary
sizes for each language are truncated at 30,000. The embedding size
and RNN hidden size are set to 256. The non-linear layers are of sizes
512 and 1,024 respectively (i.e. $b_{g}^{{\rm S}},b_{g}^{{\rm T}}\in\mathbb{R}^{512}$
and $b_{u}\in\mathbb{R}^{1024}$). Dropout probability 0.1 is applied
to word embeddings. The mini-batch size is 128. The optimizer is Adam
with learning rate $5\times10^{-4}$, and early stopping is employed
based on accuracy on the development set.

\subsection{Evaluation Metric}

We report \emph{acceptability detection accuracy} on the test set
as our evaluation metric. Note that, true positives (TP) and true
negatives (TN) are both important for our purpose because failing
to catch unacceptable translations or presenting acceptable cases
to human (thereby increasing the cost) are both undesirable. Therefore,
F score is less suitable here than e.g. information retrieval where
true negatives are innumerable. Besides, if the downstream task is
binary classification, we also present a formula to estimate the \emph{cross-lingual
accuracy} with \emph{downstream system accuracy} and \emph{acceptability
detection accuracy}.

Let $p_{f}\triangleq P\left(c=1|{\bf s}\right)$ be the probability
of predicting the correct final label given the source sentence ${\bf s}$.
Introducing a binary random variable $y$ with $y=1$ representing
the MT sentence is acceptable, we have
\begin{equation}
\begin{aligned} & P\left(c=1|{\bf s}\right)\\
= & \sum_{y}P\left(c=1|y,{\bf s}\right)P\left(y|{\bf s}\right)\\
= & P\left(c=1|y=1,{\bf s}\right)P\left(y=1|{\bf s}\right)+\\
 & P\left(c=1|y=0,{\bf s}\right)P\left(y=0|{\bf s}\right).
\end{aligned}
\end{equation}
Because the downstream task is binary classification, we have $P\left(c=1|y=1,{\bf s}\right)=P\left(c=0|y=0,{\bf s}\right)=1-P\left(c=1|y=0,{\bf s}\right)$.
Defining $p_{t}\triangleq P\left(c=1|y=1,{\bf s}\right)$ and $p_{d}\triangleq P\left(y=1|{\bf s}\right)$
gives
\begin{equation}
p_{f}=p_{t}p_{d}+\left(1-p_{t}\right)\left(1-p_{d}\right).
\end{equation}
The probabilities $p_{f},p_{t},p_{d}$ can be estimated by \emph{cross-lingual
accuracy}, \emph{downstream system accuracy}, and \emph{acceptability
detection accuracy}, respectively. This formula reflects the dependence
of cross-lingual accuracy on both MT and downstream system performance.
In our setting, the downstream system is fixed, while an improvement
of acceptability detection accuracy $\Delta p_{d}$ will bring an
overall improvement
\begin{equation}
\Delta p_{f}=\left(2p_{t}-1\right)\Delta p_{d}.
\end{equation}
Because $p_{t}>0.5$ for binary classification, improving acceptability
detection accuracy will always positively affect the cross-lingual
pipeline.

Finally, it is worth noting that acceptability detection accuracy
of the baseline reflects the proportion of acceptable instances on
test sets.

\begin{table*}[!t]
\begin{centering}
{\fontsize{10}{12}\selectfont%
\begin{tabular}{|c|c|c|c||c|c|c|}
\hline
MT & BLEU & detection model & training data size & subjectivity & sentiment & named entity\tabularnewline
\hline
\hline
\multirow{9}{*}{T-0.1m} & \multirow{9}{*}{19.89} & baseline & - & 65.79 & 74.30 & 66.90\tabularnewline
\cline{3-7} \cline{4-7} \cline{5-7} \cline{6-7} \cline{7-7}
 &  & \multirow{4}{*}{BiQuEst} & 0.1m & 65.87 (+\hphantom{0}0.08) & 74.82 (+0.52) & 72.56 (+\hphantom{0}5.66)\tabularnewline
\cline{4-7} \cline{5-7} \cline{6-7} \cline{7-7}
 &  &  & 0.2m & 65.89 (+\hphantom{0}0.10) & 74.94 (+0.64) & 72.62 (+\hphantom{0}5.72)\tabularnewline
\cline{4-7} \cline{5-7} \cline{6-7} \cline{7-7}
 &  &  & 0.5m & 66.01 (+\hphantom{0}0.22) & 75.05 (+0.75) & 72.80 (+\hphantom{0}5.90)\tabularnewline
\cline{4-7} \cline{5-7} \cline{6-7} \cline{7-7}
 &  &  & \hphantom{0.}1m & 65.96 (+\hphantom{0}0.17) & 75.17 (+0.87) & 72.90 (+\hphantom{0}6.00)\tabularnewline
\cline{3-7} \cline{4-7} \cline{5-7} \cline{6-7} \cline{7-7}
 &  & \multirow{4}{*}{BiRNN} & 0.1m & 70.53 (+\hphantom{0}4.74) & 78.96 (+4.66) & 86.79 (+19.89)\tabularnewline
\cline{4-7} \cline{5-7} \cline{6-7} \cline{7-7}
 &  &  & 0.2m & 72.77 (+\hphantom{0}6.98) & 80.82 (+6.52) & 89.20 (+22.30)\tabularnewline
\cline{4-7} \cline{5-7} \cline{6-7} \cline{7-7}
 &  &  & 0.5m & 75.52 (+\hphantom{0}9.73) & 82.04 (+7.74) & 90.68 (+23.78)\tabularnewline
\cline{4-7} \cline{5-7} \cline{6-7} \cline{7-7}
 &  &  & \hphantom{0.}1m & 75.93 (+10.14) & 83.29 (+8.99) & 91.86 (+24.96)\tabularnewline
\hline
\multirow{9}{*}{T-1.25m} & \multirow{9}{*}{34.77} & baseline & - & 73.01 & 77.75 & 72.22\tabularnewline
\cline{3-7} \cline{4-7} \cline{5-7} \cline{6-7} \cline{7-7}
 &  & \multirow{4}{*}{BiQuEst} & 0.1m & 73.08 (+0.07) & 78.18 (+0.43) & 75.76 (+\hphantom{0}3.54)\tabularnewline
\cline{4-7} \cline{5-7} \cline{6-7} \cline{7-7}
 &  &  & 0.2m & 73.07 (+0.06) & 78.31 (+0.56) & 75.85 (+\hphantom{0}3.63)\tabularnewline
\cline{4-7} \cline{5-7} \cline{6-7} \cline{7-7}
 &  &  & 0.5m & 73.08 (+0.07) & 78.25 (+0.50) & 75.96 (+\hphantom{0}3.74)\tabularnewline
\cline{4-7} \cline{5-7} \cline{6-7} \cline{7-7}
 &  &  & \hphantom{0.}1m & 73.13 (+0.12) & 78.24 (+0.49) & 76.09 (+\hphantom{0}3.87)\tabularnewline
\cline{3-7} \cline{4-7} \cline{5-7} \cline{6-7} \cline{7-7}
 &  & \multirow{4}{*}{BiRNN} & 0.1m & 73.43 (+0.42) & 79.81 (+2.06) & 86.03 (+13.81)\tabularnewline
\cline{4-7} \cline{5-7} \cline{6-7} \cline{7-7}
 &  &  & 0.2m & 74.35 (+1.34) & 80.87 (+3.12) & 87.24 (+15.02)\tabularnewline
\cline{4-7} \cline{5-7} \cline{6-7} \cline{7-7}
 &  &  & 0.5m & 76.02 (+3.01) & 82.32 (+4.57) & 89.41 (+17.19)\tabularnewline
\cline{4-7} \cline{5-7} \cline{6-7} \cline{7-7}
 &  &  & \hphantom{0.}1m & 77.52 (+4.51) & 83.19 (+5.44) & 90.35 (+18.13)\tabularnewline
\hline
\multirow{9}{*}{M-0.1m} & \multirow{9}{*}{20.55} & baseline & - & 68.38 & 76.22 & 70.06\tabularnewline
\cline{3-7} \cline{4-7} \cline{5-7} \cline{6-7} \cline{7-7}
 &  & \multirow{4}{*}{BiQuEst} & 0.1m & 68.88 (+0.50) & 76.59 (+0.37) & 73.50 (+\hphantom{0}3.44)\tabularnewline
\cline{4-7} \cline{5-7} \cline{6-7} \cline{7-7}
 &  &  & 0.2m & 68.92 (+0.54) & 76.80 (+0.58) & 73.55 (+\hphantom{0}3.49)\tabularnewline
\cline{4-7} \cline{5-7} \cline{6-7} \cline{7-7}
 &  &  & 0.5m & 69.03 (+0.65) & 76.87 (+0.65) & 73.85 (+\hphantom{0}3.79)\tabularnewline
\cline{4-7} \cline{5-7} \cline{6-7} \cline{7-7}
 &  &  & \hphantom{0.}1m & 69.18 (+0.80) & 76.94 (+0.72) & 73.81 (+\hphantom{0}3.75)\tabularnewline
\cline{3-7} \cline{4-7} \cline{5-7} \cline{6-7} \cline{7-7}
 &  & \multirow{4}{*}{BiRNN} & 0.1m & 71.30 (+2.92) & 79.46 (+3.24) & 86.57 (+16.51)\tabularnewline
\cline{4-7} \cline{5-7} \cline{6-7} \cline{7-7}
 &  &  & 0.2m & 73.19 (+4.81) & 80.62 (+4.40) & 88.12 (+18.06)\tabularnewline
\cline{4-7} \cline{5-7} \cline{6-7} \cline{7-7}
 &  &  & 0.5m & 75.75 (+7.37) & 81.73 (+5.51) & 90.08 (+20.02)\tabularnewline
\cline{4-7} \cline{5-7} \cline{6-7} \cline{7-7}
 &  &  & \hphantom{0.}1m & 76.82 (+8.44) & 82.90 (+6.68) & 90.49 (+20.43)\tabularnewline
\hline
\multirow{9}{*}{M-1.25m} & \multirow{9}{*}{24.08} & baseline & - & 71.01 & 77.06 & 71.90\tabularnewline
\cline{3-7} \cline{4-7} \cline{5-7} \cline{6-7} \cline{7-7}
 &  & \multirow{4}{*}{BiQuEst} & 0.1m & 71.13 (+0.12) & 77.44 (+0.38) & 74.71 (+\hphantom{0}2.81)\tabularnewline
\cline{4-7} \cline{5-7} \cline{6-7} \cline{7-7}
 &  &  & 0.2m & 71.17 (+0.16) & 77.45 (+0.39) & 74.65 (+\hphantom{0}2.75)\tabularnewline
\cline{4-7} \cline{5-7} \cline{6-7} \cline{7-7}
 &  &  & 0.5m & 71.36 (+0.35) & 77.58 (+0.52) & 74.92 (+\hphantom{0}3.02)\tabularnewline
\cline{4-7} \cline{5-7} \cline{6-7} \cline{7-7}
 &  &  & \hphantom{0.}1m & 71.46 (+0.45) & 77.72 (+0.66) & 75.05 (+\hphantom{0}3.15)\tabularnewline
\cline{3-7} \cline{4-7} \cline{5-7} \cline{6-7} \cline{7-7}
 &  & \multirow{4}{*}{BiRNN} & 0.1m & 72.68 (+1.67) & 79.15 (+2.09) & 86.18 (+14.28)\tabularnewline
\cline{4-7} \cline{5-7} \cline{6-7} \cline{7-7}
 &  &  & 0.2m & 73.12 (+2.11) & 80.35 (+3.29) & 87.51 (+15.61)\tabularnewline
\cline{4-7} \cline{5-7} \cline{6-7} \cline{7-7}
 &  &  & 0.5m & 76.51 (+5.50) & 81.83 (+4.77) & 88.98 (+17.08)\tabularnewline
\cline{4-7} \cline{5-7} \cline{6-7} \cline{7-7}
 &  &  & \hphantom{0.}1m & 77.85 (+6.84) & 82.41 (+5.35) & 89.88 (+17.98)\tabularnewline
\hline
\end{tabular}}
\par\end{centering}
\caption{\label{tab:transformer-100k}Acceptability detection accuracy (\%)
for different downstream tasks and MT systems (T stands for Transformer
and M stands for Moses). The BLEU scores are case-insensitive BLEU-4
calculated on NIST 2008. Absolute improvements over the baseline are
shown in parentheses.}
\end{table*}

\subsection{Downstream Tasks and Systems}

We experiment with three downstream tasks: subjectivity classification,
sentiment classification, and named entity recognition. These tasks
are framed as binary classification, three-way classification (positive,
negative, neutral), and structured prediction, respectively. Unlike
classification tasks that return a single label, named entity recognition
tags the input sentence with named entities. The exact definition
of acceptability in Equation (\ref{eq:2}) can be designed for specific
needs. In our experiments, we take acceptability to represent whether
the multisets of named entities in the reference sentence and the
MT sentence are the same.

We use off-the-shelf toolkits to perform downstream tasks: Datumbox\footnote{https://github.com/datumbox/datumbox-framework}
for subjectivity and sentiment classification, and Stanford NER\footnote{https://nlp.stanford.edu/software/CRF-NER.shtml}
for named entity recognition.

\begin{table}
\begin{centering}
\begin{tabular}{|c||c|c|c|}
\hline
 & subjectivity & sentiment & named entity\tabularnewline
\hline
\hline
TP & 5,676 & 6,764 & 6,345\tabularnewline
\hline
FP & 1,504 & 1,005 & \hphantom{0,}469\tabularnewline
\hline
TN & 1,917 & 1,565 & 2,841\tabularnewline
\hline
FN & \hphantom{0,}903 & \hphantom{0,}666 & \hphantom{0,}345\tabularnewline
\hline
\end{tabular}
\par\end{centering}
\caption{\label{tab:detailed-evaluation}Detailed evaluation of BiRNN trained
with 1m data for Transformer trained on 0.1m parallel sentence pairs.}
\end{table}

\subsection{Machine Translation Models}

Our acceptability detection framework can be applied to any translation
system as long as we can use it to perform decoding. To facilitate
experiment, we build machine translation systems of our own. We investigate
neural machine translation and phrase-based statistical machine translation,
which are the Transformer implemented in THUMT \cite{zhang_thumt:_2017}
and the Moses toolkit \cite{koehn_moses:_2007}, respectively. The
parallel corpus for training comes from LDC with 1.25m sentence pairs.
In order to vary the translation performance and represent different
training conditions, we also train with a subset of 0.1m sentence
pairs. This results in four translation models. In the following section,
we will report acceptability detection accuracy for each of them.

\section{Experimental Results}

\subsection{Performance of Acceptability Detection}

\begin{figure}
\begin{centering}
\includegraphics{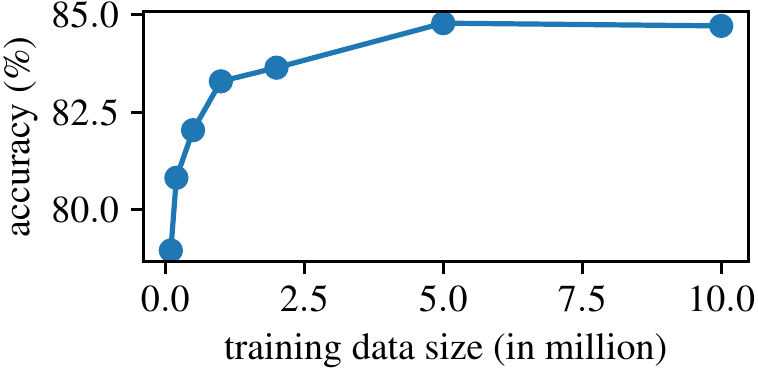}
\par\end{centering}
\caption{\label{fig:data-size}Training BiRNN on different data sizes impacts
acceptability detection accuracy. The setting is sentiment acceptability
detection for Transformer trained on 0.1m parallel sentence pairs.}
\end{figure}

We can see from the accuracy scores of acceptability detection in
Table \ref{tab:transformer-100k} that both BiQuEst and BiRNN are
able to improve over the baseline, and BiRNN performs much better
than BiQuEst. The same thing can be said for different downstream
tasks, but the degree of improvement for named entity recognition
is much larger than the other two tasks, with an absolute improvement
of 25\% for the best-performing BiRNN model. This indicates that named
entity translation issues are well captured by our models.

We also report detailed evaluation of the BiRNN-1m acceptability detection
model for Transformer-0.1m in Table \ref{tab:detailed-evaluation},
taking positive to mean acceptable. For unacceptable translations
that will be processed by the handler module, we note that TN \textgreater{}
FN. This means the majority of detected-as-unacceptable translations
are truly unacceptable (i.e. the labels they receive from the downstream
system would be different than their references). Therefore, when
the downstream task is binary classification (subjectivity classification),
flipping their labels is a better strategy than passing them untouched
as in the original cross-lingual pipeline.

\subsection{Effect of the Acceptability Training Data Size}

One advantage of our framework is the automatic generation of acceptability
annotation data from large-scale parallel corpora. It is expected
that more acceptability annotation data can result in better acceptability
detection models. This is confirmed by the accuracy scores in Table
\ref{tab:transformer-100k}: For Transformer-0.1m, BiRNN trained on
1m data improves on its 0.1m version by around 5 points, and the improvements
are similarly large for other MT systems. For BiQuEst, however, the
gain from large training set is marginal, likely due to underfitting
with the small feature set.

We further explored even more data for sentiment acceptability detection
of Transformer-0.1m, and generated up to 10m training instances. BiRNN
was also trained on subsets of 2m and 5m. The results are summarized
along with smaller-scale experiments in Figure \ref{fig:data-size}.
Increasing the training data size can bring further improvement, but
the return gradually diminishes. We also tried doubling the embedding
size and hidden size to train on 10m data, but no gain was observed
from this larger network. This may indicate that data at the 10m scale
expose little extra information for the learning of acceptability
detection task in the current experiments.

\subsection{Behavior of Different Translation Models}

Enlarging parallel corpus for the training of the translation model
will improve its translation quality; we thus obtain a Transformer
model trained on the full 1.25m parallel corpus and detect acceptability
for it, with results given in the Transformer-1.25m row of Table \ref{tab:transformer-100k}.
We first notice that the baseline accuracy gets higher for every downstream
task. This means a translation model with higher quality (as measured
by BLEU \cite{papineni_bleu:_2002}) also produces more acceptable
translations for downstream tasks. As for the acceptability detection
models, their accuracy scores remain at similar levels, although their
improvements on the baseline appear smaller.

Similar observations can be made for Moses translation models in Table
\ref{tab:transformer-100k}. In line with received wisdom, Moses can
outperform Transformer when parallel data is limited, but is superseded
when the corpus is sufficiently large. This is not only reflected
in BLEU scores, but also in the relative levels of the baseline accuracy
for acceptability detection. BiRNN models again show consistently
high performance, which indicates they are general for working with
different types of translation models.

As the baseline acceptability detection accuracies show the label
distributions of test sets, which are similar for training sets in
our experiments, it is clear that stronger MT systems will result
in more unbalanced label distributions that might hinder the learning
of acceptability detectors. However, our experiments for balancing
strategies did not yield improvement, which indicates that the current
unbalance levels are tolerable. Nevertheless, we anticipate that highly
strong MT systems diminish the need for acceptability detection, and
our approach is more beneficial for weak MT systems by helping to
close their gap with strong MT systems for downstream tasks, as the
acceptability detection accuracy improvements over the baseline (shown
in parentheses in Table \ref{tab:transformer-100k}) are consistently
larger for weaker MT systems.

\section{Discussion}

Our experiments demonstrate the utility of the acceptability detection
framework for three downstream tasks. We have seen that the named
entity recognition task is somewhat special due to its structured
output, which calls for specific decision for what exactly constitutes
$f^{{\rm T}}\left({\bf t}\right)=f^{{\rm T}}\left({\bf r}\right)$
in Equation (\ref{eq:2}). This also means other variations are possible,
for example taking $f^{{\rm T}}$ to be the number of named entities,
thereby excusing named entity translation error as long as they are
not over or under translated, if this is actually needed by the translation
quality specification. The flexibility in interpreting Equation (\ref{eq:2})
also allows combining several downstream tasks by taking $f^{{\rm T}}$
to perform multiple tasks together and return a tuple of task outputs;
this can also be equivalently seen as a single downstream task composed
of several subtasks.

Despite the flexibility, our framework is not without limitation,
as not every downstream task can fit into it. If the input to the
downstream task is more than one sentence, for example semantic textual
similarity that takes a pair of sentences, or document classification
that operates on documents, then our framework is less useful because
parallel data in those forms are rare. If the output of the downstream
task is even more complex than named entity recognition that we have
dealt with, for example summarization that outputs a sentence, then
the definition in Equation (\ref{eq:2}) would not be easily settled.
This can be attributed to the root cause that NLP tasks with unstructured
output are difficult to evaluate, even if references are provided.

Although NLP tasks with complex output are unsuitable for downstream
tasks, they open up opportunities for extending our framework to define
and detect acceptability for them. Like machine translation, their
output may also be further processed by downstream tasks. This is
especially true for automatic speech recognition and semantic parsing,
which also attract research on quality estimation for them \cite{zamani_reference-free_2015,dong_confidence_2018}.
Our framework can also build acceptability detectors for these systems
as long as abundant parallel data (parallel in the sense specific
for each complex-output task) are available.

\section{Related Work}

Machine translation quality is difficult to measure, even if we involve
human experts. The idea of introducing downstream tasks has been explored
earlier \cite{white_task-oriented_1998,jones_ilr-based_2007} by asking
humans to perform downstream tasks like reading comprehension based
on machine translated text, and the machine translation quality is
measured by human performance of those tasks. The involvement of human
means this approach cannot scale up, and is purely for the purpose
of evaluation.

Quality estimation shares similar purpose with our work. As we mention
in Section \ref{sec:Introduction}, the usual formulation of sentence-level
quality estimation outputs scores in the {[}0, 1{]} interval, leaving
users to interpret the implications. Researchers are aware that binary
labels are the most intuitive for users. A straightforward idea is
to threshold real-valued scores to obtain binary-labeled data \cite{blatz_confidence_2004,quirk_training_2004,specia_machine_2010}.
One work more related to ours \cite{turchi_data-driven_2014} proposes
an automatic method to obtain binary-labeled data that discriminate
whether the translation is useful for manual post-editing. Like usual
QE, that work is also targeted at the scenario of machine translation
for post-editing. This also means data collection is difficult due
to the high cost of human post-editing, and thus only small-scale
experiments were conducted.

Our work can also be viewed as estimating specific aspects of machine
translation quality designated by downstream tasks. In neural machine
translation, under translation and over translation are often observed
as quality issues. Focusing on these certain types of translation
failures allows researchers to design targeted detection algorithms
\cite{zheng_testing_2018}.

Besides detecting aspects of quality issues for machine translation,
researchers also attempt to make machine translation models more robust
and reduce the production of certain errors \cite{cheng_towards_2018}.
There are also endeavors to build MT systems that maintain particular
aspects of the source text like sentiment \cite{lohar_maintaining_2017,lohar_balancing_2018}.

\section{Conclusion and Future Work}

In this work, we investigate machine translation quality by proposing
a new definition called acceptability. It is binary, thus easy to
interpret. It considers downstream usage of machine translation, thus
adaptable to different quality needs. It enables large-scale annotation
by exploiting parallel data and automatic downstream systems, thus
allowing the creation of useful acceptability detectors for downstream
tasks with machine learning.

Our framework is general and encompasses various downstream tasks,
leaving much room for extension. We are also interested in exploring
more powerful detection models given the large amount of labeled data.
Finally, it may be possible to propagate labeled information to translation
models to make them aware of downstream needs.

\bibliographystyle{acl_natbib}
\bibliography{arxiv20-zm}

\end{document}